\documentclass{article}
\usepackage[T1]{fontenc}
\usepackage{iclr2027_conference}
\usepackage{times}
\usepackage{amsmath,amssymb,amsthm,mathtools}
\allowdisplaybreaks
\usepackage{booktabs,graphicx,xcolor,microtype,tikz,needspace}
\usepackage{hyperref}
\usepackage{lean_equation_links}

\definecolor{leangray}{gray}{0.35}
\newcommand{\leanref}[3]{\href{\leanbase/#1\#L#2}{\textcolor{leangray}{\texttt{#3}}}}

\newcounter{leanlinkno}
\newcommand{\lean}[2]{\stepcounter{leanlinkno}\leanref{#1}{#2}{Lean\_\theleanlinkno}}
\newcommand{\leandecl}[3]{\lean{#1}{#2}}

\makeatletter
\newcommand{\DeclareSourceEquation}[2]{\expandafter\def\csname paper@lean@#1\endcsname{#2}}
\makeatother
\newcommand{\leanbase}{https://github.com/JayPiZimmermann/Max\_Caterpillar/blob/main}
\DeclareLeanEquation{eq:width}{MaxCaterpillar/Theorems.lean}{18}
\DeclareLeanEquation{eq:count}{MaxCaterpillar/Definitions.lean}{51}
\DeclareLeanEquation{eq:word}{MaxNN/Constructions/Paramid/Caterpillar/Corrections/Formula.lean}{130}
\DeclareLeanEquation{eq:faces}{MaxNN/Constructions/Paramid/Caterpillar/Corrections/Keys.lean}{29}
\DeclareLeanEquation{eq:face-description}{MaxNN/Constructions/Paramid/Caterpillar/Corrections/Keys.lean}{29}
\DeclareLeanEquation{eq:coefficients}{MaxNN/Constructions/Paramid/Caterpillar/Corrections/Canonical/Coefficients.lean}{49}
\DeclareLeanEquation{eq:dictionary-span}{MaxCaterpillar/Lemmas.lean}{54}
\DeclareLeanEquation{eq:class-identity}{MaxCaterpillar/Lemmas.lean}{48}
\DeclareLeanEquation{stp:class}{MaxCaterpillar/Lemmas.lean}{48}
\DeclareLeanEquation{eq:geometric-class}{MaxNN/Constructions/Paramid/Geom/Faces/ConeOnly.lean}{71}
\DeclareLeanEquation{stp:corrections}{MaxNN/Constructions/Paramid/Geom/Faces/ConeOnly.lean}{71}
\DeclareLeanEquation{eq:coordinate-deletion}{MaxCaterpillar/Lemmas.lean}{41}
\DeclareLeanEquation{eq:insertion-weight-germs}{MaxNN/Constructions/Paramid/Caterpillar/Coordinate/Structured.lean}{67}
\DeclareLeanEquation{eq:graph}{MaxNN/Constructions/Paramid/Caterpillar/Corrections/Spine/GraphKey.lean}{214}
\DeclareLeanEquation{eq:cone-count}{MaxCaterpillar/Lemmas.lean}{66}
\DeclareLeanEquation{eq:features}{MaxNN/Constructions/Paramid/Caterpillar/Corrections/CompactFeatures.lean}{63}
\DeclareLeanEquation{eq:canonical-expansion}{MaxNN/Constructions/Paramid/Caterpillar/Corrections/Canonical/Coefficients.lean}{72}
\DeclareLeanEquation{eq:compilation-widths}{MaxCaterpillar/Lemmas.lean}{72}
\DeclareLeanEquation{stp:size-network}{MaxCaterpillar/Lemmas.lean}{72}
\DeclareLeanEquation{eq:asymptotics}{MaxCaterpillar/Theorems.lean}{28}
\DeclareLeanEquation{eq:network-parameters}{MaxCaterpillar/Definitions.lean}{26}
\DeclareLeanEquation{eq:network-evaluation}{MaxCaterpillar/Definitions.lean}{35}
\DeclareLeanEquation{eq:maximum}{MaxCaterpillar/Definitions.lean}{42}
\DeclareLeanEquation{eq:signed-word-sum}{MaxNN/Constructions/Paramid/Caterpillar/Formula/Values.lean}{31}
\DeclareSourceEquation{eq:hinging-maps}{https://doi.org/10.1109/TIT.2005.859246}
\DeclareSourceEquation{eq:hinging-representation}{https://doi.org/10.1109/TIT.2005.859246}
\DeclareLeanEquation{eq:max-hinge}{MaxNN/Core/Relu.lean}{24}
\DeclareLeanEquation{eq:caterpillar-data}{MaxNN/Constructions/Paramid/Caterpillar/Corrections/Formula.lean}{24}
\DeclareLeanEquation{eq:word-parent}{MaxNN/Constructions/Paramid/Caterpillar/Words.lean}{28}
\DeclareLeanEquation{eq:word-gamma}{MaxNN/Constructions/Paramid/Caterpillar/Corrections/Formula.lean}{122}
\DeclareLeanEquation{eq:edge-functions}{MaxNN/Constructions/Paramid/Caterpillar/Corrections/Formula.lean}{126}
\DeclareLeanEquation{eq:edge-comparison}{MaxNN/Constructions/Paramid/Caterpillar/Corrections/Formula.lean}{126}
\DeclareLeanEquation{eq:aux_functions}{MaxNN/Constructions/MaxN/Width/Layer/Features.lean}{75}
\DeclareLeanEquation{eq:bases}{MaxNN/Constructions/Paramid/Caterpillar/Corrections/Canonical/Average.lean}{26}
\DeclareLeanEquation{eq:gram-matrix}{MaxNN/Constructions/Paramid/Caterpillar/Corrections/Canonical/Gram.lean}{147}
\DeclareLeanEquation{eq:gram-rhs}{MaxNN/Constructions/Paramid/Caterpillar/Corrections/Canonical/Gram.lean}{201}
\DeclareLeanEquation{eq:inner-product}{MaxNN/Constructions/Paramid/Caterpillar/Corrections/Canonical/Gram.lean}{50}
\DeclareLeanEquation{eq:basis-expansion}{MaxNN/Constructions/Paramid/Caterpillar/Corrections/Canonical/Average.lean}{61}
\DeclareLeanEquation{eq:normal-equations}{MaxNN/Constructions/Paramid/Caterpillar/Corrections/Canonical/Gram.lean}{220}
\DeclareLeanEquation{eq:average-expansion}{MaxNN/Constructions/Paramid/Caterpillar/Corrections/Canonical/Coefficients.lean}{53}
\DeclareLeanEquation{eq:hinge-reversal}{MaxNN/Core/Relu.lean}{33}
\DeclareLeanEquation{eq:remainder}{MaxNN/Constructions/Paramid/Caterpillar/Corrections/AbsorbHinges.lean}{95}
\DeclareLeanEquation{eq:remainder-translation}{MaxNN/Constructions/Paramid/Caterpillar/Corrections/CompactCovariance.lean}{74}
\DeclareLeanEquation{eq:simplex}{MaxNN/Constructions/Paramid/Geom/Defs.lean}{121}
\DeclareLeanEquation{eq:support-function}{MaxNN/Constructions/Paramid/Geom/Defs.lean}{43}
\DeclareLeanEquation{eq:unit-polytope}{MaxNN/Constructions/Paramid/Geom/Boundary/ConeFaces.lean}{400}
\DeclareLeanEquation{eq:face-normal}{MaxNN/Constructions/Paramid/Geom/Defs.lean}{62}
\DeclareLeanEquation{eq:exposed-face}{MaxNN/Constructions/Paramid/Geom/Defs.lean}{62}
\DeclareLeanEquation{eq:facet-difference}{MaxNN/Constructions/Paramid/Geom/Germ.lean}{460}
\DeclareLeanEquation{eq:face-derivative}{MaxNN/Constructions/Paramid/Geom/Germ.lean}{364}
\DeclareLeanEquation{eq:facet-error}{MaxNN/Constructions/Paramid/Caterpillar/Corrections/Formula.lean}{184}
\DeclareLeanEquation{eq:linear-facet}{MaxNN/Constructions/Paramid/Caterpillar/Formula/Braid.lean}{96}
\DeclareLeanEquation{eq:lifted-induction}{MaxNN/Constructions/Paramid/Caterpillar/Formula/Lift.lean}{40}
\DeclareLeanEquation{eq:induction-error-split}{MaxNN/Constructions/Paramid/Caterpillar/Formula/Assembly.lean}{69}
\DeclareLeanEquation{eq:root-star}{MaxNN/Constructions/Paramid/Caterpillar/Coordinate/Star.lean}{25}
\DeclareLeanEquation{eq:boolean-cancellation}{MaxNN/Constructions/Paramid/Caterpillar/Words/RootStar.lean}{47}
\DeclareLeanEquation{eq:full-coordinates}{MaxNN/Constructions/Paramid/Geom/Coordinates.lean}{27}
\DeclareLeanEquation{eq:head-count-bound}{MaxCaterpillar/Asymptotics.lean}{31}
\DeclareLeanEquation{eq:cone-growth}{MaxCaterpillar/Asymptotics.lean}{44}
\DeclareLeanEquation{eq:second-growth}{MaxCaterpillar/Asymptotics.lean}{54}
\DeclareLeanEquation{eq:first-growth}{MaxCaterpillar/Asymptotics.lean}{17}
\DeclareLeanEquation{eq:absorbed-pairs}{MaxNN/Constructions/Paramid/Caterpillar/Corrections/AbsorbHinges.lean}{26}
\DeclareLeanEquation{eq:recover-coordinates}{MaxNN/Constructions/Paramid/Caterpillar/Corrections/CompactFeatures.lean}{111}
\DeclareLeanEquation{eq:feature-count}{MaxNN/Constructions/Paramid/Caterpillar/Corrections/CompactFeatures.lean}{21}
\DeclareLeanEquation{eq:normalized-description}{MaxNN/Constructions/Paramid/Caterpillar/Corrections/Normalize.lean}{24}
\DeclareLeanEquation{eq:facet-reduction}{MaxCaterpillar/Lemmas.lean}{33}
\DeclareLeanEquation{stp:reconstruction}{MaxCaterpillar/Lemmas.lean}{33}
\DeclareLeanEquation{eq:base-unit}{MaxNN/Constructions/Paramid/Caterpillar/Coordinate/Base.lean}{78}
\DeclareLeanEquation{eq:base-formula}{MaxNN/Constructions/Paramid/Caterpillar/Coordinate/Base.lean}{84}
\DeclareLeanEquation{eq:simplex-facet}{MaxNN/Constructions/Paramid/Caterpillar/Formula/SimplexGerms.lean}{60}
\DeclareLeanEquation{eq:kuhn-simplices}{MaxNN/Constructions/Paramid/Geom/Kuhn.lean}{43}
\DeclareLeanEquation{eq:kuhn-support-error}{MaxNN/Constructions/Paramid/Geom/Kuhn.lean}{320}
\DeclareLeanEquation{eq:apex-maximum}{MaxNN/Constructions/Paramid/Caterpillar/Geometry/Paramid.lean}{43}
\DeclareLeanEquation{eq:retained-count}{MaxNN/Constructions/Paramid/Caterpillar/Corrections/HingeCount.lean}{166}
\DeclareLeanEquation{stp:size-count}{MaxNN/Constructions/Paramid/Caterpillar/Corrections/HingeCount.lean}{166}
\DeclareLeanEquation{eq:face-dimension-drop}{MaxNN/Constructions/Paramid/Geom/Flag.lean}{200}
\DeclareLeanEquation{eq:error-peeling}{MaxNN/Constructions/Paramid/Geom/Faces/Peel.lean}{38}
\DeclareLeanEquation{eq:simplex-vertex-removal}{MaxNN/Constructions/Paramid/Geom/Faces/ConeOnly/SimplexGerm.lean}{184}
\DeclareLeanEquation{eq:initial-simplex-error}{MaxNN/Constructions/Paramid/Geom/Faces/ConeOnly/LocalIdentity.lean}{205}
\DeclareLeanEquation{eq:simplex-deletion-sign}{MaxNN/Constructions/Paramid/Geom/Faces/ConeOnly/SimplexGerm.lean}{195}
\DeclareLeanEquation{eq:simplex-induction-use}{MaxNN/Constructions/Paramid/Geom/Faces/ConeOnly.lean}{40}
\DeclareLeanEquation{eq:residual-invariant}{MaxNN/Constructions/Paramid/Geom/Faces.lean}{69}
\DeclareLeanEquation{eq:support-cut}{MaxNN/Constructions/Paramid/Geom/Sweep/Support.lean}{142}
\DeclareLeanEquation{eq:sweep-support-high}{MaxNN/Constructions/Paramid/Geom/Sweep.lean}{187}
\DeclareLeanEquation{eq:sweep-support-translate}{MaxNN/Constructions/Paramid/Geom/Sweep.lean}{304}
\DeclareLeanEquation{eq:sweep-operation}{MaxNN/Constructions/Paramid/Geom/Defs.lean}{106}
\DeclareLeanEquation{eq:sweep-dimension}{MaxNN/Constructions/Paramid/Geom/Sweep.lean}{366}
\DeclareLeanEquation{eq:simplex-reconstruction}{MaxNN/Constructions/Paramid/Geom/Reconstruct.lean}{83}
\DeclareLeanEquation{eq:simplex-isolate-facet}{MaxNN/Constructions/Paramid/Geom/Reconstruct/FacetGerm.lean}{268}
\DeclareLeanEquation{eq:normalized-facet-normal}{MaxNN/Constructions/Paramid/Geom/Reconstruct/FacetGerm.lean}{185}
\DeclareLeanEquation{eq:facet-group-bound}{MaxNN/Constructions/Paramid/Geom/Reconstruct/FacetGerm.lean}{326}
\DeclareLeanEquation{eq:single-facet-group}{MaxNN/Constructions/Paramid/Geom/Reconstruct/FacetGerm.lean}{326}
\DeclareLeanEquation{eq:translation-invariance}{MaxNN/Constructions/Paramid/Caterpillar/Corrections/CompactCanonical.lean}{22}
\DeclareLeanEquation{eq:tag-family}{MaxNN/Constructions/Paramid/Caterpillar/Corrections/Canonical/Table.lean}{93}
\DeclareLeanEquation{eq:tagged-coefficients}{MaxNN/Constructions/Paramid/Caterpillar/Corrections/Canonical/Coefficients.lean}{93}
\DeclareLeanEquation{eq:tagged-expansion}{MaxNN/Constructions/Paramid/Caterpillar/Corrections/Canonical/Coefficients.lean}{110}
\DeclareLeanEquation{eq:normalized-tag-coefficients}{MaxNN/Constructions/Paramid/Caterpillar/Corrections/Canonical/Table.lean}{166}
\DeclareLeanEquation{eq:active-normalized-descriptions}{MaxNN/Constructions/Paramid/Caterpillar/Corrections/Canonical/Table.lean}{171}
\DeclareLeanEquation{eq:retained-descriptions}{MaxNN/Constructions/Paramid/Caterpillar/Corrections/AbsorbHinges.lean}{31}
\DeclareLeanEquation{eq:ridge-cut-residual}{MaxNN/Constructions/Paramid/Geom/Reconstruct/SimplexCut.lean}{613}
\DeclareLeanEquation{eq:simplex-support-expansion}{MaxNN/Constructions/Paramid/Caterpillar/Formula/Completeness.lean}{80}
\DeclareLeanEquation{eq:direction-pattern}{MaxNN/Constructions/Paramid/Caterpillar/Geometry/Facets/Pattern.lean}{64}
\DeclareLeanEquation{eq:braid-reflection}{MaxNN/Constructions/Paramid/Caterpillar/Formula/Braid.lean}{66}
\DeclareLeanEquation{eq:braid-germ}{MaxNN/Constructions/Paramid/Geom/Germ/Reflection.lean}{34}
\DeclareLeanEquation{eq:block-pairing}{MaxNN/Constructions/Paramid/Caterpillar/Geometry/Toggle.lean}{42}
\DeclareLeanEquation{eq:lower-arity-translation}{MaxNN/Constructions/Paramid/Caterpillar/Formula/Lift.lean}{30}
\DeclareLeanEquation{eq:coordinate-active}{MaxNN/Constructions/Paramid/Caterpillar/Coordinate/Active.lean}{20}
\DeclareLeanEquation{eq:coordinate-fibre-partition}{MaxNN/Constructions/Paramid/Caterpillar/Coordinate/ReindexActive.lean}{228}
\DeclareLeanEquation{eq:root-star-average}{MaxNN/Constructions/Paramid/Caterpillar/Coordinate/Star.lean}{315}
\DeclareLeanEquation{eq:lower-caterpillar-summand}{MaxNN/Constructions/Paramid/Caterpillar/Coordinate/Structured.lean}{23}
\DeclareLeanEquation{eq:auxiliary-remainder}{MaxNN/Constructions/Paramid/Caterpillar/Corrections/CompactCovariance.lean}{22}
\DeclareLeanEquation{eq:compact-remainder}{MaxNN/Constructions/Paramid/Caterpillar/Corrections/CompactCovariance.lean}{89}
\DeclareLeanEquation{eq:network-assembly}{MaxCaterpillar/Construction.lean}{39}
\DeclareLeanEquation{eq:signed-remainder}{MaxNN/Constructions/Paramid/Caterpillar/Corrections/CompactCanonical.lean}{57}
\DeclareLeanEquation{eq:coordinate-relabel}{MaxNN/Constructions/Paramid/Caterpillar/Formula/CoordinateDirections.lean}{119}
\makeatletter
\AtBeginDocument{%
\let\paper@linked@tagform\tagform@
\def\tagform@#1{%
\ifpaper@displaytag
\ifmeasuring@\else
\ifcsname paper@lean@\paper@equationlabel\endcsname\else
\PackageError{caterpillar}{Missing source link for equation `\paper@equationlabel'}{Run scripts/paper_links.py and copy the updated main.tex.}%
\fi
\fi
\fi
\paper@linked@tagform{#1}%
}%
}
\makeatother

\newtheorem{theorem}{Theorem}
\newtheorem*{representationtheorem}{Theorem}
\newtheorem*{depthconjecture}{Historical conjecture}
\newtheorem{lemma}[theorem]{Lemma}
\theoremstyle{definition}
\newtheorem{definition}[theorem]{Definition}
\usepackage{refcount}
\newcommand{\pfnum}{??}
\newcounter{pfclaim}\newcounter{pfstep}
\newcounter{pfblock}
\renewcommand{\thepfclaim}{\pfnum.\arabic{pfclaim}}
\renewcommand{\thepfstep}{\pfnum.\arabic{pfstep}}

\newenvironment{Proof}[2]{%
  \stepcounter{pfblock}%
  \setcounter{pfclaim}{0}\setcounter{pfstep}{0}%
  \xdef\pfnum{\getrefnumber{#2}}%
  \begin{proof}[\normalfont\bfseries\phantomsection\label{pf:#2}Proof of #1~\ref{#2}]%
}{\end{proof}}

\newcommand{\pfstep}[2]{\par\medskip\refstepcounter{pfstep}%
  \noindent\textbf{\LeanLinkIfDeclared{#1}{Step \thepfstep}\ (#2).}\label{#1}\enspace\ignorespaces}
\newcommand{\finishes}[2]{This finishes the proof of #1~\ref{#2}.}
 {\par}
 {\end{proof}}

\iclrfinalcopy
\newcommand{\R}{\mathbb R}
\newcommand{\maxn}{\operatorname{max}}
\newcommand{\relu}{\sigma}
\newcommand{\spn}{\operatorname{span}}
\newcommand{\LS}{\mathcal L}
\newcommand{\D}{\mathcal D}
\newcommand{\B}{\mathcal B}
\newcommand{\one}{\mathbf 1}
\title{\resizebox{\textwidth}{!}{TWO HIDDEN LAYERS SUFFICE: THE CATERPILLAR NET}}
\author{Jakob Paul Zimmermann\\
\normalfont FU Berlin, TU Berlin\\
\href{mailto:jakob.paul.zimmermann@campus.tu-berlin.de}{jakob.paul.zimmermann@campus.tu-berlin.de}\\
\url{https://jakob-paul-zimmermann.com}}
\date{}
\hypersetup{hidelinks,pdftitle={TWO HIDDEN LAYERS SUFFICE: THE CATERPILLAR NET},
  pdfauthor={Jakob Paul Zimmermann}}

\begin{document}
\maketitle
\lhead{} 
\renewcommand{\headrulewidth}{0pt}
\begin{abstract}
Every continuous piecewise linear (CPWL) function is a finite signed sum of maxima of affine functions, making maximum representation a central problem in the study of the expressiveness of ReLU networks. A now-disproved conjecture predicted logarithmic depth for maxima. Inspired by the proof by \citet{BSY} that two hidden layers suffice for general CPWL functions, we give the \emph{Caterpillar Net}, a nonrecursive construction for the maximum of $N$ inputs with two hidden layers of widths $O(N^2)$ and $2^{O(N\log N)}$. The construction has a direct geometric interpretation.\footnote{Our results were obtained through proof search in Lean using Claude Code and Codex. The Lean code is available in the accompanying repository \href{https://github.com/JayPiZimmermann/Max_Caterpillar}{https://github.com/JayPiZimmermann/Max\_Caterpillar}. Links to the Lean formalization appear in theorem titles and equation numbers; equation numbers are clickable.}
\end{abstract}

\section{Introduction}

In this paper we investigate the representation capabilities of ReLU networks: What functions are representable by such networks of different depth and widths? As ReLU neural networks represent continuous piecewise-linear (CPWL) functions, a sharp question is: given any CPWL function, what is the ReLU network of smallest depth or smallest width representing this function?
The maximum is a universal building block for such CPWL functions. Write
\begin{align}\label{eq:maximum}
 \maxn_N(x):=\max_{0\leq i<N}x_i.
\end{align}

\begin{representationtheorem}[{\citealp{WangSun}}]
Let $d\geq1$ and let $f:\R^d\to\R$ be continuous and affine on each member
of a finite polyhedral cover. There are $R\in\mathbb N$, affine maps and signs
\begin{align}\label{eq:hinging-maps}
 A_t:\R^d\longrightarrow\R^{d+1},\qquad
 \varepsilon_t\in\{-1,1\}\qquad(1\leq t\leq R)
\end{align}
such that
\begin{align}\label{eq:hinging-representation}
 f(x)=\sum_{t=1}^R\varepsilon_t\maxn_{d+1}(A_t(x))
 \qquad\text{for every }x\in\R^d.
\end{align}
\end{representationtheorem}

Thus a two-hidden-layer representation of $\maxn_{d+1}$ gives one for every
CPWL function on $\R^d$: precompose with the affine maps and add the networks
in parallel.

\begin{depthconjecture}[{\citealp[Conjecture~1.5]{HertrichEtAl}, now disproved}]
For every integer $k\geq1$, the function $x\mapsto\max\{0,x_1,\ldots,x_{2^k}\}$ cannot be represented by a ReLU network with $k$ hidden layers, regardless of their widths.
\end{depthconjecture}
The case $k=2$ asserts impossibility with two hidden layers. \citet{BakaevEtAl} disproved it by constructing a two-hidden-layer representation of the maximum of five inputs.

This reduction made the geometry of the maximum a central approach to the
representation problem well before the work of \citet{BSY}.
Polyhedral approaches to depth include the Newton-polytope framework of \citet{HertrichEtAl}, the simplex subdivisions of \citet{BakaevEtAl}, and the depth lower bounds for braid-fan-compatible networks of \citet{GrilloEtAl}.
For width, \citet{SafranDepth} used extremal graph theory to prove lower bounds for exact maximum representation, while \citet{LuClosure} studied first-layer width thresholds for approximation and exact representation.
\citet{BSY} then constructed two-hidden-layer ReLU representations for
arbitrary CPWL functions in every dimension.

\begin{theorem}[\citeauthor{BSY}, \citeyear{BSY}]
 \label{thm:bsy}
    Every continuous piecewise affine function   $f:\mathbb{R}^d\to\mathbb{R}$ with finitely many affine pieces can be represented exactly by a ReLU network with two hidden layers.
\end{theorem}

Our contribution is to specialize their general construction to $\maxn_N$
and simplify it into the \emph{Caterpillar Net}. Taking $2^{\Theta(N^2)}$ as the second-layer width baseline
via applying the \citet{BSY} recursive construction, we pursue two aims:
\begin{enumerate}
\item Give a nonrecursive construction for $\maxn_N$ with a direct
geometric interpretation.
\item Reduce its second hidden-layer width.
\end{enumerate}

\Needspace{8\baselineskip}
Write $\relu(t):=\max(0,t)$ applied coordinatewise to vectors.
A network of widths $(h_1,h_2)$ has parameters
\begin{align}\label{eq:network-parameters}
 \begin{aligned}
 W_1&\in\R^{h_1\times N},& b_1&\in\R^{h_1},\\
 W_2&\in\R^{h_2\times h_1},& b_2&\in\R^{h_2},\\
 w_3&\in\R^{h_2},& b_3&\in\R,
 \end{aligned}
\end{align}

and evaluates as the function
\begin{align}\label{eq:network-evaluation}
 x\longmapsto w_3^\top\relu\bigl(W_2\relu(W_1x+b_1)+b_2\bigr)+b_3.
\end{align}
The output is affine, there are no skip connections, and equality is required
on all of $\R^N$.

\Needspace{20\baselineskip}
\begin{theorem}[Caterpillar Net, \leandecl{MaxCaterpillar/Theorems.lean}{18}{maximum_width}, \leandecl{MaxCaterpillar/Lemmas.lean}{80}{construction_width}, {\hyperref[pf:thm:main]{proof}}]\label{thm:main}
For every $N\geq3$, define
\begin{align}\label{eq:count}
 C_N=\sum_{m=2}^{N-1}\binom Nm\frac{m!}{2}(N-m)m^{N-m-1}.
\end{align}
There is a two-hidden-layer ReLU network computing $\maxn_N$ with widths
\begin{align}\label{eq:width}
 h_1=\binom N2+N+1,\qquad
 h_2=3^{N-2}C_N-N(N-1)+2.
\end{align}
In particular, these widths satisfy, as $N\to\infty$,
\begin{align}\label{eq:asymptotics}
 h_1=O(N^2),\qquad h_2=2^{O(N\log N)}.
\end{align}
\end{theorem}

\noindent\textit{Remark (small arities).} The linked theorem also covers $N=1$ and $N=2$ by separate networks of widths $(2,2)$ and $(4,2)$. At $N=3$, formula~\eqref{eq:width} gives widths $(7,5)$.

\section{The Caterpillar Net}

Two one-line identities drive the construction. For reals $s,t$ and every
linear function $\ell$ on $\R^N$,
\begin{align}
 \max(s,t)&=t+\relu(s-t),\label{eq:max-hinge}\\
 \relu(\ell(x))&=\relu(-\ell(x))+\ell(x).\label{eq:hinge-reversal}
\end{align}
The first suggests computing the maximum along a path with attached leaves.
Alternating sums of support functions of caterpillar polytopes give the correct support-function class; support functions of their proper faces, together with linear functions, correct the remaining error to pointwise equality.
The second reverses an inner hinge at the cost of a linear term; it merges equal units in Section~\ref{sec:size}.

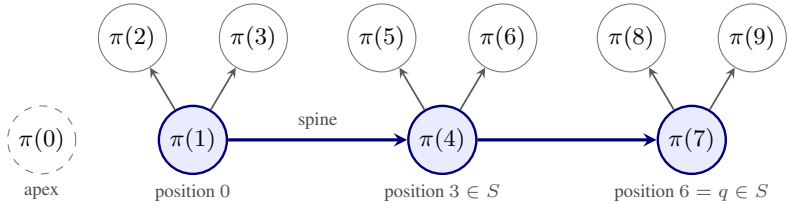
\begin{figure}[htbp]
\centering
\begin{tikzpicture}[
 x=1cm,y=1cm,>=stealth,
 every node/.style={font=\small},
 vertex/.style={circle,draw=black!55,fill=white,minimum size=9mm,inner sep=1pt},
 spine vertex/.style={vertex,draw=blue!45!black,fill=blue!8,line width=.9pt},
 leaf edge/.style={->,draw=black!65,line width=.65pt},
 spine edge/.style={->,draw=blue!45!black,line width=1.2pt},
 annotation/.style={font=\scriptsize,text=black!70}
]
\node[vertex,dashed] (apex) at (-2,0) {$\pi(0)$};
\node[annotation] at (-2,-.7) {apex};

\node[spine vertex] (h0) at (0,0) {$\pi(1)$};
\node[spine vertex] (h3) at (3.3,0) {$\pi(4)$};
\node[spine vertex] (h6) at (6.6,0) {$\pi(7)$};
\node[annotation] at (0,-.7) {position $0$};
\node[annotation] at (3.3,-.7) {position $3\in S$};
\node[annotation] at (6.6,-.7) {position $6=q\in S$};

\node[vertex] (l1) at (-.8,1.35) {$\pi(2)$};
\node[vertex] (l2) at (.8,1.35) {$\pi(3)$};
\node[vertex] (l4) at (2.5,1.35) {$\pi(5)$};
\node[vertex] (l5) at (4.1,1.35) {$\pi(6)$};
\node[vertex] (l7) at (5.8,1.35) {$\pi(8)$};
\node[vertex] (l8) at (7.4,1.35) {$\pi(9)$};

\draw[spine edge] (h0) -- node[above,annotation] {spine} (h3);
\draw[spine edge] (h3) -- (h6);
\draw[leaf edge] (h0) -- (l1);
\draw[leaf edge] (h0) -- (l2);
\draw[leaf edge] (h3) -- (l4);
\draw[leaf edge] (h3) -- (l5);
\draw[leaf edge] (h6) -- (l7);
\draw[leaf edge] (h6) -- (l8);
\end{tikzpicture}
\caption{A labelled caterpillar with $N=10$ and $S=\{3,6\}$. The spine
$\pi(1)\to\pi(4)\to\pi(7)$ has two leaves at each vertex;
all arrows point from parent to child. The apex $\pi(0)$ is separate
from the tree. The last spine position is $q=6$, so the head function
is $h_{\pi,S}(x)=x_{\pi(7)}-x_{\pi(0)}$.}
\label{fig:caterpillar}
\end{figure}

\Needspace{10\baselineskip}
Fix $N\geq3$ and write $x=(x_0,\ldots,x_{N-1})\in\R^N$.
\begin{definition}[Labelled caterpillar, \leandecl{MaxNN/Constructions/Paramid/Caterpillar/Corrections/Formula.lean}{24}{WordIndex}]\label{def:caterpillar}
A \emph{caterpillar description} is a permutation--set pair
\begin{align}\label{eq:caterpillar-data}
 \pi\in\operatorname{Sym}\{0,\ldots,N-1\},\qquad
 \varnothing\ne S\subseteq\{1,\ldots,r\},\qquad r:=N-2.
\end{align}
Position $j$ carries the label $\pi(j+1)$, for $0\leq j\leq r$, and the label $\pi(0)$ is the \emph{apex}.
The \emph{spine positions} are $\{0\}\cup S$. We also call $S$ the \emph{cut set}: a cut at $j\in S$ separates positions $j-1$ and $j$ in the ordered list. The last spine position, and the \emph{parent} of a position $j\geq1$, the nearest spine position before it, are
\begin{align}\label{eq:word-parent}
 q:=\max S,\qquad
 p(j):=\max\bigl(\{0\}\cup\{i\in S:i<j\}\bigr)\qquad(1\leq j\leq r).
\end{align}
Joining each position $j\geq1$ to its parent gives a directed tree on the labels $\pi(1),\ldots,\pi(N-1)$, the \emph{labelled caterpillar}: its spine is the path through the spine positions, and every other label is a leaf attached to the preceding spine vertex (Figure~\ref{fig:caterpillar}). The apex is separate from this tree.
\end{definition}

\begin{definition}[Caterpillar function and face functions, \leandecl{MaxNN/Constructions/Paramid/Caterpillar/Corrections/Formula.lean}{130}{rawCone}, \leandecl{MaxNN/Constructions/Paramid/Caterpillar/Corrections/Keys.lean}{29}{faceKey}]\label{def:caterpillar-function}\label{def:face-functions}
For labels $i,j$, for the tree edges, and for the head, put
\begin{align}
 \ell_{ij}(x)&:=x_i-x_j,\label{eq:edge-comparison}\\
 L_{\pi,S}&:=\bigl\{\ell_{\pi(p(j)+1),\pi(j+1)}:1\leq j\leq r\bigr\},\label{eq:edge-functions}\\
 h_{\pi,S}&:=\ell_{\pi(q+1),\pi(0)}.\label{eq:word-gamma}
\end{align}
The \emph{edge functions} $L_{\pi,S}$ subtract each child's coordinate from its parent's, one for each of the $r$ tree edges, and the \emph{head function} $h_{\pi,S}$ compares the last spine vertex with the apex.
For a linear function $h$, a finite set $L$ of linear functions, and a \emph{state} $\tau(\ell)\in\{0,\ \ell,\ \relu\circ\ell\}$ chosen for every $\ell\in L$, define
\begin{align}
 U^{\tau}_{h,L}(x)&:=\relu\Bigl(h(x)+\sum_{\ell\in L}\tau(\ell)(x)\Bigr).\label{eq:faces}
\end{align}
Thus each $\ell\in L$ is omitted, kept linearly, or kept as a hinge. Write $U_{h,L}:=U^{\tau}_{h,L}$ for the all-hinge state $\tau(\ell)=\relu\circ\ell$. We call $(h,L)$ a \emph{function description} of $U_{h,L}$: it records the linear head and the inner hinges. The caterpillar description $(\pi,S)$ specifies the labelled tree from which such functions are generated.
The \emph{caterpillar function} of $(\pi,S)$ and its \emph{face functions} are
\begin{align}\label{eq:word}
 U_{\pi,S}:=U_{h_{\pi,S},L_{\pi,S}},\qquad
 U^{\tau}_{\pi,S}:=U^{\tau}_{h_{\pi,S},L_{\pi,S}}.
\end{align}
\end{definition}

The term \emph{face function} comes from the following geometry.
Identify linear functions with their coefficient vectors. For a polytope $P$, its \emph{support function}, and the polytope whose support function is $U_{h,L}$, are
\begin{align}
 h_P(x)&:=\max_{p\in P}\langle p,x\rangle,\label{eq:support-function}\\
 Q_{h,L}&:=\operatorname{conv}\Bigl(\{0\}\cup\Bigl(h+\sum_{\ell\in L}[0,\ell]\Bigr)\Bigr),\qquad
 U_{h,L}=h_{Q_{h,L}}.\label{eq:unit-polytope}
\end{align}
Indeed, the segment $[0,\ell]$ has support function $\relu\circ\ell$, Minkowski sums add support functions, and adjoining $0$ applies the outer ReLU.
The head and the $N-2$ edge functions are jointly linearly independent: their label edges form a tree on all $N$ labels, since the head connects the apex to the last spine vertex. Thus the base is an $(N-2)$-dimensional parallelotope whose affine hull does not contain $0$, and $Q_{h,L}$ is an $(N-1)$-dimensional pyramid.
A face of $Q_{h,L}$ containing the apex $0$ and another point is the convex hull of $0$ and a face of the base. Since the base is a translated sum of segments, the face exposed by a linear functional is obtained by maximizing on each segment separately, which selects the endpoint $0$, the endpoint $\ell$, or the whole segment. These have support functions $0$, $\ell$, and $\relu\circ\ell$, and the outer ReLU adjoins the apex. Hence the face functions~\eqref{eq:faces} include the support function of every face of $Q_{h,L}$ through the apex.

\Needspace{8\baselineskip}
\begin{definition}[Dictionary, \leandecl{MaxCaterpillar/Construction.lean}{68}{dictionary}]\label{def:dictionary}
The \emph{dictionary} $\D$ is the finite set of functions $\R^N\to\R$ consisting of every face function $U^{\tau}_{\pi,S}$, over all caterpillar descriptions $(\pi,S)$ and all states $\tau$, together with the \emph{auxiliary functions}
\begin{align}\label{eq:aux_functions}
 \begin{aligned}
 &x\mapsto\relu(x_i),\quad x\mapsto\relu(-x_i)
 &&(0\leq i<N),\\
 &x\mapsto\relu(x_i-x_j)
 &&(0\leq i<j<N).
 \end{aligned}
\end{align}
\end{definition}
Heads and edge functions are coordinate differences, so face functions and pair hinges are translation invariant: with $\one:=(1,\ldots,1)\in\R^N$,
\begin{align}\label{eq:translation-invariance}
 U^{\tau}_{\pi,S}(x+t\one)=U^{\tau}_{\pi,S}(x)\qquad(t\in\R).
\end{align}
Only the coordinate features $\relu(\pm x_i)$ change under $x\mapsto x+t\one$.
Theorem~\ref{thm:dictionary} below proves $\maxn_N\in\spn\D$; the following rule then fixes one real coefficient for each dictionary function.

To turn coefficients of distinct functions into literal network data, retain the finite tags used to generate the dictionary. Let $\mathcal A=\mathcal A_{\mathrm{face}}\sqcup\mathcal A_{\mathrm{aux}}$, where $\mathcal A_{\mathrm{face}}$ contains the distinct head--edge-set pairs obtained by collecting each linear state into the head as in~\eqref{eq:face-description}, and $\mathcal A_{\mathrm{aux}}$ contains the indexed auxiliary features in~\eqref{eq:aux_functions}. If $v(a)$ is the function represented by a tag, then
\begin{align}\label{eq:tag-family}
 v:\mathcal A\longrightarrow\D,\qquad \D=v(\mathcal A).
\end{align}
Different tags may represent the same function.

\paragraph[An exact coefficient rule]{An exact coefficient rule (\leandecl{MaxNN/Constructions/Paramid/Caterpillar/Corrections/Canonical/Gram.lean}{172}{gramMatrix_isUnit})}
Let
\begin{align}\label{eq:bases}
 \B:=\{S\subseteq\D:S\text{ is a basis of }\spn\D\}.
\end{align}
This set is nonempty because the finite spanning set $\D$ contains a basis.
For $S\in\B$, form
\begin{align}\label{eq:gram-matrix}
 G_S&:=\bigl(\langle g,h\rangle\bigr)_{g,h\in S},\\
 b_S&:=\bigl(\langle g,\maxn_N\rangle\bigr)_{g\in S}.\label{eq:gram-rhs}
\end{align}
using the inner product
\begin{align}\label{eq:inner-product}
 \langle f,g\rangle:=\int_{[-1,1]^N}f(x)g(x)\,dx.
\end{align}
Every dictionary function is continuous and positively homogeneous. If a linear combination of the basis functions has zero squared integral, continuity makes it zero throughout the cube, and positive rescaling then makes it zero on all of $\R^N$. Linear independence forces every coefficient to vanish, so $G_S$ is positive definite and invertible. Write $G_S^{-1}$ for its inverse matrix.
For each $g\in\D$, average over the bases containing $g$:
\begin{align}\label{eq:coefficients}
 c_g:=\frac{1}{|\B|}
         \sum_{\substack{S\in\B\\g\in S}}(G_S^{-1}b_S)_{g}.
\end{align}
\begin{lemma}[Canonical expansion, \leandecl{MaxNN/Constructions/Paramid/Caterpillar/Corrections/Canonical/Coefficients.lean}{72}{sum_canonicalCoeff_eq}]\label{lem:coefficients}
Assuming Theorem~\ref{thm:dictionary}, the coefficients~\eqref{eq:coefficients} satisfy
\begin{align}\label{eq:canonical-expansion}
 \maxn_N=\sum_{g\in\D}c_gg.
\end{align}
\end{lemma}
\begin{Proof}{Lemma}{lem:coefficients}
Fix $S\in\B$. By Theorem~\ref{thm:dictionary}, there is $d\in\R^S$ such that
\begin{align}\label{eq:basis-expansion}
 \maxn_N=\sum_{g\in S}d_g g.
\end{align}
Taking the inner product with each $h\in S$ gives
\begin{align}\label{eq:normal-equations}
 \begin{aligned}
 (b_S)_h&=\sum_{g\in S}\langle h,g\rangle d_g=(G_Sd)_h,\\
 d&=G_S^{-1}b_S.
 \end{aligned}
\end{align}
Interchanging the finite sums in~\eqref{eq:coefficients} therefore yields
\begin{align}\label{eq:average-expansion}
 \sum_{g\in\D}c_gg
 =\frac{1}{|\B|}\sum_{S\in\B}\sum_{g\in S}(G_S^{-1}b_S)_g\,g
 =\frac{1}{|\B|}\sum_{S\in\B}\maxn_N
 =\maxn_N.
\end{align}
\finishes{Lemma}{lem:coefficients}
\end{Proof}
For each tag, split the coefficient of its represented function equally over its nonempty finite fiber:
\begin{align}\label{eq:tagged-coefficients}
 \widetilde c_a:=\frac{c_{v(a)}}{|\{b\in\mathcal A:v(b)=v(a)\}|}.
\end{align}
While computationally intractable to extract explicit weights for large $N$, averaging over all bases avoids arbitrary symmetry-breaking and provides a canonical coefficient map ideal for formal verification.
Regrouping by the fibers of $v$ gives the literal tagged expansion
\begin{align}\label{eq:tagged-expansion}
 \maxn_N=\sum_{a\in\mathcal A}\widetilde c_a v(a).
\end{align}
Thus duplicate face and auxiliary tags neither duplicate nor lose any coefficient.

\section{Why the dictionary spans the maximum}

The maximum $\maxn_N$ is the support function of $\operatorname{conv}\{e_0,\ldots,e_{N-1}\}$. To work with a full dimensional polytope, let us do the following substitution: for
\begin{align}\label{eq:simplex}
 \Delta:=\operatorname{conv}\{0,e_1,\ldots,e_{N-1}\}\subseteq H_0:=\{x\in\R^N:x_0=0\}
\end{align}
we observe for an arbitrary input $x\in\R^N$ that
\begin{align}\label{eq:full-coordinates}
 \maxn_N(x)=x_0+h_\Delta(x-x_0\one).
\end{align}
To prove that $\maxn_N\in\spn\D$ it suffices to show that $h_\Delta$ lies in the span of the restrictions of $\D$ to $H_0$. Indeed, $x_0=\relu(x_0)-\relu(-x_0)$ lies in the span of the auxiliary functions~\eqref{eq:aux_functions}, and by~\eqref{eq:translation-invariance} the substitution $x\mapsto x-x_0\one$ fixes every dictionary function except the coordinate features $\relu(\pm x_i)$, which become $\relu(\pm(x_i-x_0))$; by~\eqref{eq:hinge-reversal} these are linear combinations of auxiliary functions. Hence $x\mapsto g(x-x_0\one)$ belongs to $\spn\D$ for every $g\in\D$, and~\eqref{eq:full-coordinates} gives the claimed implication.

All polytopes in the geometric argument below lie in the $(N-1)$-dimensional space $H_0$. We restrict heads and edge functions to $H_0$ and identify them with their coefficient vectors there. Let $\LS_k$ denote the real span of support functions of polytopes in $H_0$ of affine dimension at most $k$. In particular, $\LS_0$ is the space of linear functions, since the support function of a point $p$ is $x\mapsto\langle p,x\rangle$.
\Needspace{8\baselineskip}
For a direction
\begin{align}\label{eq:face-normal}
u\in H_0,
\end{align}
the exposed face of a polytope $P\subseteq H_0$ with normal $u$ is the set of points of $P$ maximizing
the linear function $p\mapsto\langle p,u\rangle$:
\begin{align}\label{eq:exposed-face}
 P^u:=\{p\in P:\langle p,u\rangle=h_P(u)\}.
\end{align}
For $u\ne0$, it is where the supporting hyperplane with outward normal $u$ meets $P$. The opposite face $P^{-u}$ consists of the minimizers; for $u=0$, both faces are $P$ itself.
Define the function
\begin{align}\label{eq:facet-difference}
 D_uh_P(x):=h_{P^u}(x)-h_{P^{-u}}(x)
 \qquad(x\in H_0).
\end{align}
Here $u$ selects the two faces, while $x$ is the input to their support functions. In particular, $D_0=0$.
Geometrically, $D_u$ acts as a derivative that replaces the support function of a polytope by the difference of the support functions of its opposite exposed faces. If $\langle u,\cdot\rangle$ is nonconstant on $P$, these are proper boundary faces of smaller dimension; if it is constant, the two support functions cancel. This removes the full-dimensional contribution and explains the dimension drop in~\eqref{eq:face-dimension-drop}. For a simplex and a facet normal $u$, $D_u$ is a \emph{facet-isolating operator}: it returns that facet's support function minus the linear support function of the opposite vertex. This is the mechanism used in Step~\ref{stp:reconstruction}, rather than a local gradient calculation.

\Needspace{10\baselineskip}
For a finite linear combination $g$ of support functions, define its one-sided directional derivative at $u$ by
\begin{align}\label{eq:face-derivative}
\begin{aligned}
\partial_u g(x)&:=\lim_{t\downarrow0}\frac{g(u+tx)-g(u)}{t},\\
\partial_u h_P&=h_{P^u},\qquad D_ug=\partial_ug-\partial_{-u}g.
\end{aligned}
\end{align}
Thus the map $h_P\mapsto h_{P^u}$ extends linearly to linear combinations of support functions, independently of their decomposition. If $\dim P\leq k$, either $\langle u,\cdot\rangle$ is constant on $P$, so $D_uh_P=0$, or both selected faces have dimension at most $k-1$. Consequently,
\begin{align}\label{eq:face-dimension-drop}
D_u\LS_k\subseteq\LS_{k-1}\qquad(k\geq1).
\end{align}

\paragraph{The signed caterpillar sum.}
We seek a combination that agrees with $h_\Delta$ up to a function in $\LS_{N-2}$.
Adding $x_{\pi(0)}$ makes the caterpillar function the following maximum
\begin{align}\label{eq:apex-maximum}
  x_{\pi(0)}+U_{\pi,S}(x)=\max\left\{x_{\pi(0)},\,x_{\pi(q+1)}+\sum_{\ell\in L_{\pi,S}}\sigma(\ell(x))\right\}.
\end{align}
Averaging over permutations gives
the required symmetry
\begin{align}\label{eq:signed-word-sum}
 F_N(x):=\frac{1}{N!}\sum_{\pi}
   \sum_{\varnothing\ne S\subseteq\{1,\ldots,r\}}
   (-1)^{N+|S|}\bigl(x_{\pi(0)}+U_{\pi,S}(x)\bigr),
\end{align}
where the signs $(-1)^{N+|S|}$ are chosen to cancel the extra contributions from support functions of exposed caterpillar faces, modulo $\LS_{N-3}$. Establishing the required cancellations is a main step of the proof below.

\Needspace{8\baselineskip}
\begin{theorem}[Dictionary sufficiency, \leandecl{MaxCaterpillar/Lemmas.lean}{54}{maximum_mem_dictionary}, {\hyperref[pf:thm:dictionary]{proof}}]\label{thm:dictionary}
For every $N\geq3$, the dictionary of Definition~\ref{def:dictionary}
satisfies
\begin{align}\label{eq:dictionary-span}
 \maxn_N\in\spn\D.
\end{align}
\end{theorem}

\begin{Proof}{Theorem}{thm:dictionary}
Recall from this section's introduction that it suffices to show that $h_\Delta$ lies in the span of the restrictions of $\mathcal D$ to $H_0$.

Define the error by
\begin{align}\label{eq:facet-error}
\begin{aligned}
f:H_0&\longrightarrow\R,\\
f(x)&:=F_N(x)-h_\Delta(x)\\
&=F_N(0,x_1,\ldots,x_{N-1})-\max\{0,x_1,\ldots,x_{N-1}\}.
\end{aligned}
\end{align}
Step~\ref{stp:corrections} reduces dictionary membership to $f\in\LS_{N-2}$. Step~\ref{stp:reconstruction} shows that it suffices to prove $D_uf\in\LS_{N-3}$ for every direction $u$. Step~\ref{stp:class} proves these bounds by induction on $N$: dimension bounds and cancellation handle all directions except those with one coordinate different from all the others, which share a common value. In that case, deleting the exceptional coordinate invokes the induction hypothesis.

\pfstep{stp:corrections}{Reduce to a lower-dimensional error}
In this step we reduce the problem to proving
\begin{align}\label{eq:class-identity}
  f=F_N-h_\Delta\in\LS_{N-2}.
\end{align}
Assume this relation temporarily, and write $Q_{\pi,S}:=Q_{h_{\pi,S},L_{\pi,S}}$, so that $U_{\pi,S}=h_{Q_{\pi,S}}$ by~\eqref{eq:unit-polytope}. Since the apex terms in~\eqref{eq:signed-word-sum} are linear, the assumed relation gives
\begin{align}\label{eq:geometric-class}
\frac{1}{N!}\sum_{\pi}\sum_{\varnothing\ne S\subseteq\{1,\ldots,N-2\}}(-1)^{N+|S|}h_{Q_{\pi,S}}-h_\Delta\in\LS_{N-2}.
\end{align}

We will correct this error by differentiating its support-function terms via $\partial_u$ to lower the residual dimension until the residual is linear. The resulting support functions of proper faces of the $Q_{\pi,S}$ already lie in the dictionary span: the support function of the convex hull of the apex with a base face is a face function from Definition~\ref{def:face-functions}, given by~\eqref{eq:faces}, while the support function of a base face is a linear function plus inner hinges. Coordinate features represent the linear function, and~\eqref{eq:hinge-reversal} puts either orientation of each hinge in the span of~\eqref{eq:aux_functions}. The apex alone has zero support function, and $h_{Q_{\pi,S}}=U_{\pi,S}$ is itself a dictionary function. Differentiating the simplex support-function term also produces support functions of proper simplex faces. Their membership in the dictionary span is supplied by induction on the number of vertices.

\paragraph{Claim (induction on $|T|$).}
For every nonempty $T\subseteq\{0,e_1,\ldots,e_{N-1}\}$, the support function of $\Delta_T:=\operatorname{conv}T$ lies in the dictionary span.

\paragraph{Start ($|T|=1$).}
The support function of a singleton is linear and therefore lies in the dictionary span.

\paragraph{Step ($|T|\geq2$).}
Fix $T$ and assume that $h_{\Delta_A}\in\spn\D$ for every nonempty vertex set $A$ with $|A|<|T|$, in particular for every nonempty proper subset of $T$. Consider a current vertex set $A\subseteq\{0,e_1,\ldots,e_{N-1}\}$ with $|A|\geq2$. To remove $v\in A$, choose a direction whose value at $v$ is strictly larger than its common value at all other vertices. The positive exposed face is then the singleton $\{v\}$, and the negative exposed face is the simplex on the remaining vertices. Writing $\Delta_A:=\operatorname{conv}A$, we have
\begin{align}\label{eq:simplex-vertex-removal}
\begin{aligned}
u_v&:=\begin{cases}-\sum_{i=1}^{N-1}e_i&v=0,\\v&v\ne0,\end{cases}\\
(\Delta_A)^{u_v}&=\{v\},\qquad (\Delta_A)^{-u_v}=\Delta_{A\setminus\{v\}},\\
D_{u_v}h_{\Delta_A}(x)&=\langle v,x\rangle-h_{\Delta_{A\setminus\{v\}}}(x).
\end{aligned}
\end{align}
Choose an ordering $v_1,\ldots,v_{N-|T|}$ of the vertices outside $T$. Denote the error associated with the remaining vertex set $T$ by
\begin{align}\label{eq:initial-simplex-error}
\varepsilon_T:=D_{u_{v_{N-|T|}}}\cdots D_{u_{v_1}}\left(\frac{1}{N!}\sum_\pi\sum_{\varnothing\ne S\subseteq\{1,\ldots,N-2\}}(-1)^{N+|S|}h_{Q_{\pi,S}}-h_\Delta\right).
\end{align}
The operators act from right to left, removing $v_1$, then $v_2$, and so on. In~\eqref{eq:simplex-vertex-removal}, the support function of the remaining simplex carries a minus sign, while the removed singleton contributes the linear function $x\mapsto\langle v,x\rangle$. Every subsequent $D_u$ annihilates linear functions. Thus the $N-|T|$ deletions give
\begin{align}\label{eq:simplex-deletion-sign}
D_{u_{v_{N-|T|}}}\cdots D_{u_{v_1}}h_\Delta=(-1)^{N-|T|}h_{\Delta_T}+\ell_T,
\end{align}
where $\ell_T$ is linear. The original error contains $-h_\Delta$, so the coefficient of $h_{\Delta_T}$ in $\varepsilon_T$ is $-(-1)^{N-|T|}=(-1)^{N-|T|+1}$. All other terms are linear functions or support functions of faces of the $Q_{\pi,S}$, since each $D_u$ replaces a support function by a difference of support functions of exposed faces.
The coefficient of $h_{\Delta_T}$ in $\varepsilon_T$ is $\pm1$, and all other terms already lie in the dictionary span. Thus proving $\varepsilon_T\in\spn\D$ completes the induction step.
To prove this membership, we use the additional information $\varepsilon_T\in\LS_{|T|-2}$, obtained by repeated application of~\eqref{eq:face-dimension-drop}. This bound starts the inner induction below: each reduction lowers the dimension bound by one while subtracting a function in the dictionary span, until the remaining error lies in $\LS_0$ and is therefore linear.
For the full vertex set, the operator composition is empty and means the identity, so $\varepsilon_T$ is exactly the error in~\eqref{eq:geometric-class}.

For this fixed $T$, define the remaining errors as functions $\varepsilon_T^{(j)}:H_0\to\R$ recursively by
\begin{align}\label{eq:error-peeling}
\begin{aligned}
\varepsilon_T^{(0)}(x)&:=\varepsilon_T(x),\\
\varepsilon_T^{(j+1)}(x)&:=\varepsilon_T^{(j)}(x)-\sum_{u\in E_{T,j}}\bigl(\partial_u\varepsilon_T^{(j)}\bigr)(x).
\end{aligned}
\end{align}
Here $x\in H_0$ and $0\leq j<|T|-2$. At stage $j$, the finite set of directions $E_{T,j}\subseteq H_0$ is chosen from the current error by the construction below. Thus $\varepsilon_T^{(j)}$ is the error left after $j$ derivative sums have been subtracted from $\varepsilon_T$; the superscript counts stages and is not a power.

\paragraph{Claim (induction on the reduction stage $j$).}
The sets $E_{T,j}$ can be chosen successively so that, for $0\leq j\leq|T|-2$,
\begin{align}\label{eq:residual-invariant}
\begin{aligned}
\varepsilon_T^{(j)}&\in\LS_{|T|-2-j},\\
\varepsilon_T-\varepsilon_T^{(j)}&\in\spn\D.
\end{aligned}
\end{align}
Each remaining error is a linear combination of $h_{\Delta_T}$, support functions of proper faces of $\Delta_T$, support functions of faces of the $Q_{\pi,S}$, and linear functions. The second condition says that the part already removed is represented by the dictionary.

\paragraph{Start ($j=0$).}
By definition, $\varepsilon_T^{(0)}=\varepsilon_T$. Its dimension bound and support-function expression were established above, and the removed part is zero.

\paragraph{Step ($j<|T|-2$).}
Assume the directions for the first $j$ stages have been chosen and the claim holds at stage $j$. Put $k:=|T|-2-j\geq1$. We require each new direction $u$ to be nonorthogonal to the simplex direction space $\operatorname{dir}\Delta_T:=\operatorname{span}\{v-w:v,w\in T\}$. Its values on $T$ are then not all equal, so the maximizing vertices form a nonempty proper subset:
\begin{align}\label{eq:simplex-induction-use}
\begin{aligned}
A_u&:=\{v\in T:\langle u,v\rangle=h_{\Delta_T}(u)\},\qquad \varnothing\ne A_u\subsetneq T,\\
\partial_u h_{\Delta_T}&=h_{\Delta_{A_u}}\in\spn\D.
\end{aligned}
\end{align}
The last membership uses the outer induction hypothesis, since $|A_u|<|T|$. An orthogonal direction would leave $h_{\Delta_T}$ unchanged and would not allow this use of the hypothesis.

Choose an auxiliary expression for $\varepsilon_T^{(j)}$ by support functions of polytopes of dimension at most $k$, and collect their distinct $k$-dimensional direction spaces. For each such space $V$, choose $u_V\in V^\perp$ not orthogonal to $\operatorname{dir}\Delta_T$ or to any other collected space $V'$. This is possible because $\dim\Delta_T=|T|-1>k$: each of $\operatorname{dir}\Delta_T$ and $V'$ contains a vector outside $V$, whose nonzero projection onto $V^\perp$ is not orthogonal to that vector. Adding sufficiently small multiples of these projections avoids each orthogonality condition in turn while preserving those already avoided. Choose $E_{T,j}:=\{u_V:V\text{ is a collected direction space}\}$, with $E_{T,j}=\varnothing$ if there are none. In the sum of derivatives over $E_{T,j}$, each term coming from a $k$-dimensional polytope is preserved exactly once. Every other derivative of that term is the support function of a proper face, while terms of lower dimension remain in $\LS_{k-1}$. Subtracting this derivative sum therefore cancels all $k$-dimensional terms. By the recursive definition~\eqref{eq:error-peeling}, the resulting error satisfies $\varepsilon_T^{(j+1)}\in\LS_{k-1}=\LS_{|T|-3-j}$.

Now compute these same derivatives using the support-function expression in the claim. The derivative of $h_{\Delta_T}$ lies in the dictionary span by~\eqref{eq:simplex-induction-use}. Differentiating the support function of an already proper simplex face still gives a support function of a proper face of $\Delta_T$, so the same induction hypothesis applies. Differentiating the support function of a face of $Q_{\pi,S}$ gives a support function of a face of the same polytope, covered by the opening dictionary description, and linear functions remain linear. Thus the derivative sum lies in the dictionary span. By~\eqref{eq:error-peeling}, $\varepsilon_T-\varepsilon_T^{(j+1)}$ is the sum of the previously removed part and this derivative sum, so it also lies in the dictionary span. Subtracting the derivative sum preserves the stated types of support-function terms. This proves the claim at stage $j+1$.

At stage $j=|T|-2$, the remaining error lies in $\LS_0$ and is linear. Together with~\eqref{eq:residual-invariant}, this gives $\varepsilon_T\in\spn\D$. Its vertex-deletion expression has coefficient $(-1)^{N-|T|+1}\in\{-1,1\}$ on $h_{\Delta_T}$ and all other terms in the dictionary span, so $h_{\Delta_T}\in\spn\D$. This completes the induction step on $|T|$. Taking $T$ to be the full vertex set proves $h_\Delta\in\spn\D$ under the temporary assumption~\eqref{eq:class-identity}.
\finishes{Step}{stp:corrections}

\pfstep{stp:reconstruction}{Why the facet bounds imply the class identity}
For $N\geq4$ and $f=F_N-h_\Delta$, we show that $f\in\LS_{N-2}$ follows from
\begin{align}\label{eq:facet-reduction}
D_uf\in\LS_{N-3}\qquad\text{for every }u\in H_0.
\end{align}
The idea is to express $f$ modulo $\LS_{N-2}$ using support functions of full-dimensional simplices. The facet bounds then force their weighted contributions to cancel modulo $\LS_{N-2}$.

The base of each $Q_{\pi,S}$ is an affine cube. Order its edge vectors by a bijection $\omega:\{1,\ldots,N-2\}\to L_{\pi,S}$, add them one at a time, and adjoin the apex:
\begin{align}\label{eq:kuhn-simplices}
K_{\pi,S,\omega}:=\operatorname{conv}\left(\{0\}\cup\left\{h_{\pi,S}+\sum_{a=1}^k\omega(a):0\leq k\leq N-2\right\}\right).
\end{align}
These $(N-2)!$ simplices form the \emph{Kuhn subdivision}: in cube coordinates, each occupies one ordering region $1\geq t_{\omega(1)}\geq\cdots\geq t_{\omega(N-2)}\geq0$. For a hyperplane cut $P=P_+\cup P_-$, the support-function identity
\begin{align}\label{eq:support-cut}
h_P+h_{P_+\cap P_-}=h_{P_+}+h_{P_-}
\end{align}
shows that subdivision changes the sum of support functions only by support functions of the cutting sections. Hence
\begin{align}\label{eq:kuhn-support-error}
h_{Q_{\pi,S}}-\sum_\omega h_{K_{\pi,S,\omega}}\in\LS_{N-2}.
\end{align}
Substituting~\eqref{eq:kuhn-support-error} into~\eqref{eq:signed-word-sum} expresses $f$ as a finite linear combination of support functions of $(N-1)$-dimensional simplices in $H_0$, plus a remainder in $\LS_{N-2}$. The simplex $\Delta$ contributes its support function with coefficient $-1$, and the linear apex terms are included in the remainder.

For a simplex $P$, an outward facet normal $\nu_F$ exposes the facet $F$ on one side and its opposite vertex $v_F$ on the other. Thus
\begin{align}\label{eq:simplex-isolate-facet}
D_{\nu_F}h_P(x)=h_F(x)-\langle v_F,x\rangle.
\end{align}
Directions exposing no facet on either side give support functions of faces of dimension at most $N-3$. Consequently,~\eqref{eq:facet-reduction} says that each weighted, signed sum of parallel-facet support functions belongs to $\LS_{N-3}$.

Our assumption controls the weighted facet contributions, whereas our conclusion concerns $f$. To recover the original sum modulo $\LS_{N-2}$, we use the same linear operator for every simplex, so cancellations between facet contributions survive reconstruction. Define
\begin{align}\label{eq:sweep-operation}
(\mathcal Tg)(x):=\begin{cases}g(x)&x_1\geq0,\\g(x-x_1e_1)&x_1<0.\end{cases}
\end{align}
The sign in this definition concerns the input $x$. To interpret it geometrically, first suppose that every point $p\in P$ satisfies $p_1\geq0$, and write $\operatorname{pr}(p):=p-p_1e_1$ for projection onto $p_1=0$. Since $\langle p,x\rangle-\langle\operatorname{pr}(p),x\rangle=p_1x_1$, each point dominates its projection when $x_1\geq0$, and its projection dominates when $x_1<0$. Consequently,
\begin{align}\label{eq:sweep-support-high}
\mathcal T h_P(x)=\max\{h_P(x),h_{\operatorname{pr}P}(x)\}=h_{\operatorname{conv}(P\cup\operatorname{pr}P)}(x).
\end{align}
This convex hull fills the segments joining points of $P$ to their projections, so its dimension is at most $\dim P+1$. For an arbitrary $P$, choose $c\geq0$ such that $P+ce_1$ lies in $p_1\geq0$. Translation adds $cx_1$ to the support function, and $\mathcal T(x_1)=\relu(x_1)$, giving
\begin{align}\label{eq:sweep-support-translate}
\mathcal T h_P(x)=h_{\operatorname{conv}((P+ce_1)\cup\operatorname{pr}(P+ce_1))}(x)-c\relu(x_1).
\end{align}
The correction $\relu(x_1)$ is the support function of the segment $[0,e_1]$. Thus linearity gives
\begin{align}\label{eq:sweep-dimension}
\mathcal T\LS_k\subseteq\LS_{k+1}.
\end{align}

\paragraph{Claim (induction on $k$).}
Every $(N-1)$-dimensional simplex $P\subseteq H_0$ with exactly $k$ facets not parallel to $e_1$ satisfies
\begin{align}\label{eq:simplex-reconstruction}
h_P-\mathcal T\left(\sum_{F\text{ facet of }P}\operatorname{sgn}\langle\nu_F,e_1\rangle\,h_F\right)\in\LS_{N-2}.
\end{align}
Here $\nu_F$ is an outward normal of $F$. Its sign against $e_1$ records whether the facet bounds lines parallel to $e_1$ from above or below; parallel facets contribute zero. Boundedness ensures both nonzero signs occur, so $k\geq2$.

\paragraph{Start ($k=2$).}
There is one upper facet $F^+$ and one lower facet $F^-$. Every line parallel to $e_1$ that meets $P$ has its two intersection endpoints on these facets. Thus $h_P(x)=h_{F^+}(x)$ when $x_1\geq0$, and $h_P(x)=h_{F^-}(x)$ when $x_1<0$. When $x_1\geq0$, $\mathcal T$ evaluates at $x$, so subtracting $\mathcal T(h_{F^+}-h_{F^-})$ from $h_P$ leaves $h_{F^-}$. When $x_1<0$, $\mathcal T$ evaluates both facet support functions at $x-x_1e_1$, where they agree because the facets have the same projection. Their difference is therefore zero, again leaving $h_{F^-}$. The reconstruction error is therefore $h_{F^-}\in\LS_{N-2}$, proving the claim for $k=2$.

\paragraph{Step ($k>2$).}
Assume the claim for $k-1$, and let $P$ have $k$ facets not parallel to $e_1$. Two of them have the same nonzero sign. Their intersection is an $(N-3)$-dimensional face $C$. Let $W:=\operatorname{span}\{p-q:p,q\in C\}$, and project orthogonally within $H_0$ onto $W^\perp$. This is a two-dimensional plane, since $\dim H_0=N-1$ and $\dim W=N-3$. All points of $C$ have the same projection $c$, because their differences lie in $W$. There are just two vertices of $P$ outside $C$, so the projected simplex is a triangle with vertices $c$ and the projections of those two vertices. The chosen facets project to its two sides meeting at $c$.

The chosen facet normals are orthogonal to $W$, so their signs against $e_1$ equal their signs against its projection. This projected direction is nonzero, since the chosen facets are not parallel to $e_1$. If both sides are upper, moving from $c$ opposite to this direction enters the triangle; if both are lower, moving in this direction enters it. Continuing meets the interior of the opposite edge. The inverse image of the line through $c$ in this direction is a hyperplane in $H_0$ containing $C$ and parallel to $e_1$. It cuts the edge joining the two vertices outside $C$ at an interior point and divides $P$ into two $(N-1)$-dimensional simplices. Each keeps one of the two chosen facets and replaces the other by the new facet parallel to $e_1$. Its other facets retain their normal directions, so each new simplex has exactly $k-1$ facets not parallel to $e_1$.

Write $E(Q)$ for the reconstruction error on the left of~\eqref{eq:simplex-reconstruction}. Let $P_+$ and $P_-$ be the two new simplices, let $R:=P_+\cap P_-$ be their common cutting facet, and write $F_1,F_2$ for the chosen facets and $R_F:=R\cap F$ for every other facet of $P$. With $s_F:=\operatorname{sgn}\langle\nu_F,e_1\rangle$, applying~\eqref{eq:support-cut} to $P$ and to every split facet gives the exact residual identity
\begin{align}\label{eq:ridge-cut-residual}
E(P)=E(P_+)+E(P_-)-h_R+\mathcal T\left(\sum_{\substack{F\text{ facet of }P\\F\notin\{F_1,F_2\}}}s_Fh_{R_F}\right).
\end{align}
Indeed, the inherited facets retain their signs, while the two copies of $R$ are parallel to $e_1$ and therefore have sign zero. By the induction hypothesis, $E(P_+),E(P_-)\in\LS_{N-2}$; moreover, $\dim R=N-2$ and $\dim R_F\leq N-3$. Equation~\eqref{eq:sweep-dimension} therefore puts every term on the right of~\eqref{eq:ridge-cut-residual} in $\LS_{N-2}$. This proves the claim for $k$ and completes the induction.

\paragraph{Conclusion for $f$.}
Apply~\eqref{eq:simplex-reconstruction} to every simplex support-function term in the expansion of $f$, keeping its original coefficient. A facet $F$ parallel to $e_1$ has $\langle\nu_F,e_1\rangle=0$, so its term in the reconstruction formula is zero and can be omitted. For each remaining occurrence of a facet, choose $\nu_F$ outward from its simplex and put
\begin{align}\label{eq:normalized-facet-normal}
u_F:=\frac{\nu_F}{\langle\nu_F,e_1\rangle},\qquad \langle u_F,e_1\rangle=1.
\end{align}
Thus the coordinate with index $1$ is normalized to $1$; the coordinate with index $0$ remains zero because $u_F\in H_0$. Oppositely oriented parallel normals give the same $u_F$. Only the normal is rescaled: the support function and its coefficient are unchanged.

Let $\mathcal U$ be the finite set of vectors $u_F$ obtained this way. For each $u\in\mathcal U$, define $G_u$ by adding all facet support-function terms with $u_F=u$, retaining their signs and weights. An occurrence of $F$ as a facet of $K_{\pi,S,\omega}$ contributes $(-1)^{N+|S|}\operatorname{sgn}\langle\nu_F,e_1\rangle\,h_F/N!$, while an occurrence as a facet of $\Delta$ contributes $-\operatorname{sgn}\langle\nu_F,e_1\rangle\,h_F$. These are precisely the terms with the same normalized normal, including all repetitions.

More explicitly, define $\rho$ to be the sum of the linear apex terms and the weighted Kuhn subdivision errors in~\eqref{eq:kuhn-support-error}. Then
\begin{align}\label{eq:simplex-support-expansion}
f=\frac1{N!}\sum_\pi\sum_{\varnothing\ne S\subseteq\{1,\ldots,N-2\}}(-1)^{N+|S|}\sum_\omega h_{K_{\pi,S,\omega}}-h_\Delta+\rho,
\qquad \rho\in\LS_{N-2}.
\end{align}
For a simplex $P$, collect its facets with normalized normal $u$ as
\begin{align}\label{eq:single-facet-group}
B_u(P):=\sum_{\substack{F\text{ facet of }P\\\langle\nu_F,e_1\rangle\ne0,\ u_F=u}}\operatorname{sgn}\langle\nu_F,e_1\rangle h_F.
\end{align}
An empty sum is zero. If $\langle\nu_F,e_1\rangle>0$, the direction $u_F$ exposes $F$; if it is negative, $-u_F$ exposes $F$. The opposite face is a single vertex, so~\eqref{eq:simplex-isolate-facet} gives the signed facet term up to a linear function. If neither $u$ nor $-u$ exposes a facet of $P$, both exposed faces have dimension at most $N-3$. Thus $D_uh_P-B_u(P)\in\LS_{N-3}$. Applying this to the full weighted expansion, with the same permutations and nonempty cut sets as in~\eqref{eq:signed-word-sum}, gives
\begin{align}\label{eq:facet-group-bound}
\begin{aligned}
D_uf-G_u&=\sum_{\pi,S\ne\varnothing}\frac{(-1)^{N+|S|}}{N!}\sum_\omega\left(D_uh_{K_{\pi,S,\omega}}-B_u(K_{\pi,S,\omega})\right)\\
&\quad-\left(D_uh_\Delta-B_u(\Delta)\right)+D_u\rho\in\LS_{N-3}
\qquad(u\in\mathcal U).
\end{aligned}
\end{align}
Together with~\eqref{eq:facet-reduction}, relation~\eqref{eq:facet-group-bound} puts every $G_u$ in $\LS_{N-3}$. By linearity, the sum of the reconstructed facet terms is $\mathcal T(\sum_{u\in\mathcal U}G_u)$, which lies in $\LS_{N-2}$ by~\eqref{eq:sweep-dimension}. Its difference from $f$ consists of the reconstruction and subdivision remainders, also in $\LS_{N-2}$. Hence $f\in\LS_{N-2}$.
\finishes{Step}{stp:reconstruction}

\Needspace{5\baselineskip}
\pfstep{stp:class}{Prove the class identity by induction}
\paragraph{Claim (induction on $N$).}
The class identity~\eqref{eq:class-identity} holds for every $N\geq3$.

\paragraph{Start ($N=3$).}
Here $r=1$, so the nonempty cut set $S$ must be $\{1\}$. The tree has the single edge $\pi(1)\to\pi(2)$, with $q=1$ and $p(1)=0$. Equation~\eqref{eq:apex-maximum} gives
\begin{align}\label{eq:base-unit}
\begin{aligned}
x_{\pi(0)}+U_{\pi,\{1\}}(x)&=\max\bigl\{x_{\pi(0)},x_{\pi(2)}+\relu(x_{\pi(1)}-x_{\pi(2)})\bigr\}\\
&=\max\{x_{\pi(0)},x_{\pi(1)},x_{\pi(2)}\}=\maxn_3(x).
\end{aligned}
\end{align}
Each term has coefficient $(-1)^{3+1}/3!=1/6$. For each apex label there are two orders of the remaining labels, and both give $\maxn_3$. Grouping the six positive terms by their apex gives
\begin{align}\label{eq:base-formula}
F_3(x)=\frac16\left(\underbrace{2\maxn_3(x)}_{\pi(0)=0}+\underbrace{2\maxn_3(x)}_{\pi(0)=1}+\underbrace{2\maxn_3(x)}_{\pi(0)=2}\right)=\maxn_3(x).
\end{align}
On $H_0$, $x_0=0$ and $\maxn_3(x)=\max\{0,x_1,x_2\}=h_\Delta(x)$. Hence $f=0\in\LS_1$, proving~\eqref{eq:class-identity} for $N=3$.

\paragraph{Step ($N\geq4$).}
Assume the class identity for $N-1$. By Step~\ref{stp:reconstruction}, it suffices to prove $D_uf\in\LS_{N-3}$ for every $u\in H_0$. By~\eqref{eq:facet-difference}, applying $D_u$ replaces each support function $h_P$ by $h_{P^u}-h_{P^{-u}}$: we select the maximizing and minimizing faces in direction $u$ and subtract their support functions. The linear apex terms disappear. For $\Delta$, these faces contain exactly the vertices with largest and smallest coordinate values $u_i$, including $u_0=0$. For a caterpillar base, the segment $[0,\ell_{ij}]$ remains free when $u_i=u_j$; its maximizing endpoint is $\ell_{ij}$ when $u_i>u_j$ and $0$ when $u_i<u_j$. The minimizing face uses the opposite choices.

We discard support-function contributions already in $\LS_{N-3}$ and cancel the remaining ones. Induction is needed only when one of $u_0,\ldots,u_{N-1}$ differs from the common value of all the others: the exposed simplex facet and the surviving caterpillar sum then give the same problem with $N-1$ inputs.

\paragraph{Discarding contributions in $\LS_{N-3}$.}
Fix $u\ne0$, since $D_0f=0$. A face of $Q_{\pi,S}$ contained in its base has a support function that is a linear function plus support functions of segments, hence belongs to $\LS_1\subseteq\LS_{N-3}$; the apex alone contributes zero. If at least two edges join labels with unequal $u_i$, at most $N-4$ segments remain free, so even adjoining the apex gives dimension at most $N-3$. Thus we need only faces containing the apex and an exposed base face, with at most one edge joining unequal values.

Removing the unique unequal edge splits the tree into two components on which the $u_i$ are constant. Treat the selected exposed sign as positive, replacing $u$ by $-u$ for the negative face and renaming the levels. The condition that the apex and exposed base tie gives
\begin{align}\label{eq:direction-pattern}
\begin{aligned}
 u_{\pi(0)}&\in\{c,d\}&&\text{for a spine edge joining levels $c,d$},\\
 u_{\pi(0)}&=c+\relu(c-d)\in\{c,2c-d\}&&\text{for a leaf of level $d$ at level $c$}.
\end{aligned}
\end{align}
Indeed, the exposed base value is $\max\{c,d\}$ in the spine case and $c+\relu(c-d)$ in the leaf case. With no unequal edge all tree values agree. Consequently, only two distinct coordinate values or the three values $c+t,c-t,c,\ldots,c$ with $t\ne0$ can contribute. This exhausts both exposed signs; all other directions are already handled by the dimension bound.

\paragraph{Cancellation for three distinct values.}
For $c+t,c-t,c,\ldots,c$, let $R$ be the reduced linear map induced by exchanging the two exceptional simplex vertices and subtracting the new value at vertex $0$. It is a rank-one reflection: for a linear functional $b$ with $\langle b,u\rangle=1$, and a linear function $\ell$, permutation symmetry gives
\begin{align}\label{eq:braid-reflection}
 R x=x-2\langle b,x\rangle u,\qquad R u=-u,\qquad F_N(Rx)=F_N(x)+\ell(x).
\end{align}
Positive homogeneity then compares the one-sided derivatives explicitly:
\begin{align}\label{eq:braid-germ}
 D_u(F_N|_{H_0})(x)=2F_N(-u)\langle b,x\rangle-\ell(x).
\end{align}
Thus
\begin{align}\label{eq:linear-facet}
D_u(F_N|_{H_0})\in\LS_0.
\end{align}
The two exposed simplex faces are vertices, so their support functions are linear as well.

\paragraph{Cancellation for two values, each repeated.}
After the preceding dimension bound, every term still requiring cancellation has a unique split
$u_{\pi(1)}=\cdots=u_{\pi(k)}\ne u_{\pi(k+1)}=\cdots=u_{\pi(r+1)}$ with $k\in S$. If the apex label has the latter value, toggle membership of $k-1$ in $S$. Otherwise, exchange the equal-valued labels $\pi(k+1),\pi(k+2)$ and toggle membership of $k+1$. Repetition of both values gives $k\geq2$ in the first case and $k+2\leq r+1$ in the second, so the operations are defined; the unchanged cut at $k$ keeps the cut set nonempty. The transformed term remains in the same case, and repeating the operation returns the original term. If $\Phi$ denotes this involution on the active terms and $w_{\pi,S}:=(-1)^{N+|S|}/N!$, then
\begin{align}\label{eq:block-pairing}
 w_{\Phi(\pi,S)}=-w_{\pi,S},\qquad
 D_u h_{Q_{\pi,S}}-D_u h_{Q_{\Phi(\pi,S)}}\in\LS_1.
\end{align}
Inactive fixed points of the global transformation are already in $\LS_{N-3}$ by the dimension bound. Summing the active pairs therefore puts $D_uF_N$ in $\LS_{N-3}$. Each exposed simplex face has at most $N-2$ vertices and hence dimension at most $N-3$, so $D_uf\in\LS_{N-3}$ without using induction.

\paragraph{One exceptional coordinate: the induction hypothesis.}
We are still working in $H_0$, so $u_0=x_0=0$. It remains to consider two values, one occurring only once. Choose a permutation $\tau$ such that $\tau(0)$ is the exceptional label. Relabeling must be followed by subtracting the new zeroth coordinate to stay in $H_0$:
\begin{align}\label{eq:coordinate-relabel}
B_\tau:H_0\longrightarrow H_0,\qquad (B_\tau x)_i:=x_{\tau(i)}-x_{\tau(0)}\qquad(0\leq i<N).
\end{align}
Its zeroth coordinate is zero, and its inverse is $B_{\tau^{-1}}$. It sends $u$ to $(0,t,\ldots,t)$ with $t\ne0$. This change preserves every $\LS_k$ and changes $f$ only by a linear term, annihilated by $D$; we transform both the input and the direction. Rescaling the direction, and reversing the sign of $D_u$ if necessary, reduces to $u=(0,1,\ldots,1)$ within $H_0$.

This direction exposes the $(N-2)$-dimensional facet $\Delta':=\operatorname{conv}\{e_1,\ldots,e_{N-1}\}$ and the opposite vertex $0$. Hence
\begin{align}\label{eq:simplex-facet}
D_uh_\Delta(x)=h_{\Delta'}(x)=\max\{x_1,\ldots,x_{N-1}\}\qquad(x\in H_0).
\end{align}
Here is the exact use of the induction hypothesis. Every caterpillar maximum increases by a common added constant. Averaging over the $(N-1)!$ lower label orders cancels the factorial in the total signed coefficient, giving
\begin{align}\label{eq:lower-arity-translation}
\begin{aligned}
&W_{N-1}:=\sum_{\varnothing\ne T\subseteq\{1,\ldots,N-3\}}(-1)^{N-1+|T|},\\
&F_{N-1}(x_1,\ldots,x_{N-1})-\max_{1\leq i\leq N-1}x_i\\
&\quad=F_{N-1}(0,x_2-x_1,\ldots,x_{N-1}-x_1)\\
&\qquad-\max\{0,x_2-x_1,\ldots,x_{N-1}-x_1\}+(W_{N-1}-1)x_1.
\end{aligned}
\end{align}
The remaining $N-1$ inputs need not start with zero. Subtracting $x_1$ puts them in the lower-arity slice $\{z\in\R^{N-1}:z_0=0\}$, exactly the domain of the induction hypothesis. Apply the $(N-1)$-input class identity to $(0,x_2-x_1,\ldots,x_{N-1}-x_1)$ on the right. Linear precomposition preserves $\LS_{N-3}$, and the last term is linear. Since $(N-1)-2=N-3$, this gives
\begin{align}\label{eq:lifted-induction}
\bigl(x\mapsto F_{N-1}(x_1,\ldots,x_{N-1})-\max\{x_1,\ldots,x_{N-1}\}\bigr)\in\LS_{N-3}.
\end{align}
Thus we must match the exposed caterpillar support functions with the lower-arity sum. We prove
\begin{align}\label{eq:coordinate-deletion}
D_{(0,1,\ldots,1)}(F_N|_{H_0}) -\bigl(x\mapsto F_{N-1}(x_1,\ldots,x_{N-1})\bigr)\in\LS_{N-3}.
\end{align}
To identify the contributing terms, put $p:=\pi^{-1}(0)\in\{0,\ldots,N-1\}$. Deleting this position gives an order $\bar\pi$ that maps positions $\{0,\ldots,N-2\}$ bijectively to labels $\{1,\ldots,N-1\}$. Modulo $\LS_{N-3}$ it is enough to retain precisely the cases
\begin{align}\label{eq:coordinate-active}
 (p=1\ \text{and}\ 1\in S),\qquad\text{or}\qquad
 p=j+2\ \text{and}\ (j+1\notin S\ \text{or}\ j=N-3),\quad 0\leq j\leq N-3.
\end{align}
The omitted cases have a base-only face, an apex alone, or an apex-base face with too few free segments, and therefore lie individually in $\LS_{N-3}$. For $0\leq i\leq N-3$, let $\iota_i$ insert a gap at the prospective pendant position $i+1$. Given a lower cut set $T\subseteq\{1,\ldots,N-3\}$, the three active insertion rules are
\begin{align}\label{eq:coordinate-fibre-partition}
\begin{aligned}
 \iota_i(t)&:=\begin{cases}t,&t<i+1,\\t+1,&t\geq i+1,\end{cases}\\
 \text{first end:}\quad &(p,S)=\left(1,\{1\}\cup\{t+1:t\in T\}\right),\\
 \text{pendant $i$:}\quad &(p,S)=\left(i+2,\{\iota_i(t):t\in T\}\right)\qquad(0\leq i\leq N-3),\\
 \text{last end:}\quad &(p,S)=\left(N-1,T\cup\{N-2\}\right).
\end{aligned}
\end{align}
In every row the upper permutation is uniquely obtained by inserting label $0$ at position $p$ in $\bar\pi$. Conversely, with $t$ restricted to $\{1,\ldots,N-3\}$, the deletion rules are $T=\{t:t+1\in S\}$ for the first end, $T=\{t:\iota_i(t)\in S\}$ for pendant $i$, and $T=S\setminus\{N-2\}$ for the last end. These operations are inverse. Criterion~\eqref{eq:coordinate-active} assigns every active index uniquely to the first row, to the pendant row when $j+1\notin S$, or to the last row when $j=N-3$ and $N-2\in S$. Thus a nonempty $T$ has $N-2$ pendant and two endpoint preimages, whereas $T=\varnothing$ has only the two endpoints because upper cut sets are nonempty.

Extend the lower caterpillar notation to $T=\varnothing$ by setting its last spine position and every parent equal to $0$, so its head begins at the root $\bar\pi(1)$. For every lower label order $\bar\pi$ put
\begin{align}\label{eq:lower-caterpillar-summand}
\begin{aligned}
C_{\bar\pi,T}(x)&:=x_{\bar\pi(0)}+U_{\bar\pi,T}(x),\\
q_T&:=\max(\{0\}\cup T),\qquad
p_T(j):=\max(\{0\}\cup\{t\in T:t<j\}).
\end{aligned}
\end{align}
For each upper preimage $(\pi,S)$ of $(\bar\pi,T)$, the insertion identities and the literal factorial coefficients are
\begin{align}\label{eq:insertion-weight-germs}
\begin{aligned}
 \text{pendant:}\quad&
 \frac{(-1)^{N+|S|}}{N!}
 =-\frac1N\frac{(-1)^{N-1+|T|}}{(N-1)!},
 &D_u\!\left(x_{\pi(0)}+U_{\pi,S}\right)+C_{\bar\pi,T}&\in\LS_1,\\
 \text{endpoint:}\quad&
 \frac{(-1)^{N+|S|}}{N!}
 =\frac1N\frac{(-1)^{N-1+|T|}}{(N-1)!},
 &D_u\!\left(x_{\pi(0)}+U_{\pi,S}\right)-C_{\bar\pi,T}&\in\LS_1.
\end{aligned}
\end{align}
Thus every one of the $N-2+2=N$ preimages contributes $1/N$ of the lower summand modulo $\LS_{N-3}$, and together they recover its coefficient in $F_{N-1}$.

The extra terms are the two endpoint fibres whose deletion leaves $T=\varnothing$, hence a rooted star. By~\eqref{eq:insertion-weight-germs}, each contributes $1/N$ of the weighted lower summand modulo $\LS_1$. For a fixed lower order $\bar\pi$, the nonroot positions are $J=\{0,2,\ldots,N-2\}$, including the original apex position $0$. For $a\in J$, let $\bar\pi^{(a)}$ exchange positions $0$ and $a$, keeping the root at position $1$; the exchange at $a=0$ is the identity. Put
\begin{align}\label{eq:root-star}
\begin{aligned}
q&:=|J|=N-2,\qquad d_a:=x_{\bar\pi(1)}-x_{\bar\pi(a)}\quad(a\in J),\\
T_A&:=\relu\left(\sum_{a\in A}d_a+\sum_{a\in J\setminus A}\relu(d_a)\right)\quad(A\subseteq J),\\
C_{\bar\pi^{(a)},\varnothing}&=x_{\bar\pi(a)}+T_{\{a\}}\quad(a\in J).
\end{aligned}
\end{align}
The Boolean alternating identity gives
\begin{align}\label{eq:boolean-cancellation}
\sum_{A\subseteq J}(-1)^{|A|}T_A=0
\quad\Longrightarrow\quad
\sum_{a\in J}T_{\{a\}}=T_\varnothing+
\sum_{\substack{A\subseteq J\\|A|\geq2}}(-1)^{|A|}T_A.
\end{align}
The empty term is a sum of support functions of segments. For $|A|\geq2$, the polytope with support function $T_A$ has one head direction and $q-|A|$ segment directions, hence dimension at most $1+q-|A|\leq q-1=N-3$. Thus the singleton sum lies in $\LS_{N-3}$; the sum of the linear apex coordinates lies in $\LS_0$. Each fixed-position exchange is a bijection on lower label orders, so
\begin{align}\label{eq:root-star-average}
 q\sum_{\bar\pi}C_{\bar\pi,\varnothing}
 =\sum_{\bar\pi}\sum_{a\in J}C_{\bar\pi^{(a)},\varnothing}\in\LS_{q-1}.
\end{align}
Since $q>0$, the full empty-cut sum belongs to $\LS_{q-1}=\LS_{N-3}$. The empty-cut coefficient $(-1)^{N-1}/(N-1)!$ and the factor $2/N$ from the two endpoint fibres are scalars, so the extra star sum lies there as well. This proves~\eqref{eq:coordinate-deletion}.

Combining~\eqref{eq:simplex-facet}, \eqref{eq:lifted-induction}, and~\eqref{eq:coordinate-deletion} now gives
\begin{align}\label{eq:induction-error-split}
\begin{aligned}
D_uf(x)={}&\bigl[D_u(F_N|_{H_0})(x)-F_{N-1}(x_1,\ldots,x_{N-1})\bigr]\\
&+\bigl[F_{N-1}(x_1,\ldots,x_{N-1})-\max\{x_1,\ldots,x_{N-1}\}\bigr].
\end{aligned}
\end{align}
The first bracket belongs to $\LS_{N-3}$ by coordinate deletion, the second by the induction hypothesis. This proves the bound for every direction. Step~\ref{stp:reconstruction} gives $f\in\LS_{N-2}$, completing the induction, and Step~\ref{stp:corrections} gives the dictionary representation.
\finishes{Theorem}{thm:dictionary}
\end{Proof}

\section{Network size}\label{sec:size}

We prove Theorem~\ref{thm:main} directly. First we bound the number of normalized caterpillar face descriptions. Then we construct one shared first layer and account for both widths.

\begin{Proof}{Theorem}{thm:main}
Fix $N\geq3$. Theorem~\ref{thm:dictionary} and Lemma~\ref{lem:coefficients}, followed by~\eqref{eq:tagged-expansion}, give the exact tagged expansion of $\maxn_N$.

\pfstep{stp:size-count}{Count the caterpillar face functions}
We first rewrite each face function using only its adjusted head and the hinges that remain. This lets us recognize when different face descriptions require the same second-layer neuron. An edge kept linearly is added to the head, an omitted edge disappears, and an edge kept as a hinge stays in the set $M$. Thus
\begin{align}\label{eq:face-description}
 U^{\tau}_{h,L}=U_{h',M},\qquad
 h':=h+\sum_{\tau(\ell)=\ell}\ell,\qquad
 M:=\{\ell\in L:\tau(\ell)=\relu\circ\ell\}.
\end{align}
The resulting function description is $(h',M)$ in the sense of Definition~\ref{def:caterpillar-function}: $h'$ is the adjusted head, and $M$ contains precisely the retained hinges. When two resulting pairs have the same adjusted head and the same hinge set, they can use one second-layer neuron, with their output coefficients added. Distinct pairs may still represent the same function, so counting these pairs gives an upper bound. The shared first layer is constructed in the next step.
We next orient the remaining hinges consistently. By~\eqref{eq:hinge-reversal}, applied to each $\ell_{ij}\in M$ with $i>j$, the same face function $U_{h',M}$ has the \emph{normalized description} $(h^+,M^+)$, where
\begin{align}\label{eq:normalized-description}
\begin{aligned}
   M^+&:=\{\ell_{\min(i,j),\max(i,j)}:\ell_{ij}\in M\},\\
   h^+&:=h'+\sum_{\ell_{ij}\in M,\ i>j}\ell_{ij}.
\end{aligned}
\end{align}
Every remaining hinge now has orientation $x_i-x_j$ with $i<j$, and the additional linear terms have been absorbed into $h^+$. No edge function of a caterpillar is the negative of another one, so $M$ and $M^+$ have the same size.

For a caterpillar description $(\pi,S)$, the spine endpoints are labelled $\pi(1)$ and $\pi(q+1)$, and every label other than $\pi(1)$ and the apex is the child of exactly one edge. Hence the normalized description of $U_{\pi,S}$ is
\begin{align}\label{eq:graph}
\begin{split}
h_{\pi,S}^+(x)&=\sum_{\ell_{ij}\in L_{\pi,S}}x_{\max(i,j)}+x_{\pi(1)}+x_{\pi(q+1)}
-\sum_{i=0}^{N-1}x_i,\\
L_{\pi,S}^+&=\{\ell_{\min(i,j),\max(i,j)}:\ell_{ij}\in L_{\pi,S}\}.
\end{split}
\end{align}
It depends only on the undirected tree, the apex, and the unordered pair of spine endpoints, not on the order of the leaves at a spine vertex or on the orientation of the spine. For a spine with $m$ vertices, choose its vertex set in $\binom Nm$ ways, its order up to reversal in $m!/2$ ways, its apex in $N-m$ ways, and the attachment of each of its $N-m-1$ leaves in $m$ ways. Reversal has no fixed point since $m\geq2$. Every caterpillar has such data, so by~\eqref{eq:count}
\begin{align}\label{eq:cone-count}
 \bigl|\{(h^+_{\pi,S},L^+_{\pi,S}):(\pi,S)\text{ a caterpillar description}\}\bigr|\leq C_N.
\end{align}

Reversing an edge function swaps its omitted and linear states, so normalizing before or after choosing the states gives the same descriptions. Every face tag therefore has a normalized description obtained from $(h^+_{\pi,S},L^+_{\pi,S})$ by choosing one of three states for each of its $N-2$ edge functions. Let $\mathcal K_0$ be the set of these normalized descriptions and let $\nu:\mathcal A_{\mathrm{face}}\to\mathcal K_0$ be normalization. For $k=(h,M)\in\mathcal K_0$, write $U_k:=U_{h,M}$. With~\eqref{eq:cone-count}, $|\mathcal K_0|\leq3^{N-2}C_N$. Only equal normalized descriptions are merged; distinct members of $\mathcal K_0$ are not identified merely because they represent the same function.

Among them are the pair-hinge descriptions $(\ell_{ij},\varnothing)$ with $i\ne j$: for the all-cut path $S=\{1,\ldots,r\}$ with apex $\pi(0)=j$ and last spine vertex $\pi(r+1)=i$, omitting every edge function gives
\begin{align}\label{eq:absorbed-pairs}
 U^{0}_{\pi,S}=\relu\circ\ell_{ij}\qquad(0\leq i\ne j<N).
\end{align}
The linear heads $\ell_{ij}$ are distinct for distinct ordered pairs. Their pair-hinge functions are distinct as well: equality of two such ReLUs makes their positive half-spaces equal, so their heads are positive scalar multiples, and equality on that half-space makes the scalar one. Let $\mathcal P$ be the subset of $\mathcal K_0$ consisting of every normalized description with empty hinge set and head equal to a coordinate difference; these are exactly the keys eligible for absorption into the first-layer remainder. The descriptions above show that $|\mathcal P|\geq N(N-1)$. Aggregate the tagged coefficients only along equal normalized face descriptions:
\begin{align}\label{eq:normalized-tag-coefficients}
 d_k:=\sum_{\substack{a\in\mathcal A_{\mathrm{face}}\\\nu(a)=k}}\widetilde c_a.
\end{align}
Discard zero fiber sums, then remove the absorbable keys:
\begin{align}\label{eq:active-normalized-descriptions}
 \mathcal K_{\mathrm{act}}:=\{k\in\mathcal K_0:d_k\ne0\}.
\end{align}
\begin{align}\label{eq:retained-descriptions}
 K:=\mathcal K_{\mathrm{act}}\setminus\mathcal P.
\end{align}
By~\eqref{eq:hinge-reversal}, every unit indexed by $\mathcal P$ lies in the span of the auxiliary functions and needs no second-layer unit. Since $K\subseteq\mathcal K_0\setminus\mathcal P$,
\begin{align}\label{eq:retained-count}
|K|\leq3^{N-2}C_N-N(N-1).
\end{align}
\finishes{Step}{stp:size-count}

\pfstep{stp:size-network}{Compile the network and establish its widths}
The terms not assigned a retained neuron form the explicitly known remainder
\begin{align}\label{eq:remainder}
R:=\sum_{a\in\mathcal A_{\mathrm{aux}}}\widetilde c_a v(a)+\sum_{k\in\mathcal K_{\mathrm{act}}\cap\mathcal P}d_kU_k.
\end{align}
By~\eqref{eq:tagged-expansion} and~\eqref{eq:normalized-tag-coefficients}--\eqref{eq:retained-descriptions}, $\maxn_N=\sum_{k\in K}d_kU_k+R$. Expand the absorbed pair hinges using~\eqref{eq:hinge-reversal} to write $R$ as a linear combination of the auxiliary functions~\eqref{eq:aux_functions}. The first layer outputs the vector
\begin{align}\label{eq:features}
\Phi(x):=\bigl(\relu(x_0),\ \relu(-x_0),\ (\relu(x_i-x_0))_{i>0},\ (\relu(x_i-x_j))_{i<j}\bigr).
\end{align}
Thus $W_1$ has rows $e_0^\top$ and $-e_0^\top$, then $e_i^\top-e_0^\top$ for $i>0$, then $e_i^\top-e_j^\top$ for $i<j$, using a fixed ordering within each block. Its row count is
\begin{align}\label{eq:feature-count}
\binom N2+(N-1)+2=\binom N2+N+1=h_1.
\end{align}
By~\eqref{eq:hinge-reversal}, these features recover all coordinates,
\begin{align}\label{eq:recover-coordinates}
\begin{aligned}
x_0&=\relu(x_0)-\relu(-x_0),\\
x_i&=x_0+\relu(x_i-x_0)-\relu(x_0-x_i)\qquad(i>0).
\end{aligned}
\end{align}
For each $k=(h,M)\in K$, substitute~\eqref{eq:recover-coordinates} into $h(x)+\sum_{\ell\in M}\relu(\ell(x))$ and collect the coefficients of $\Phi(x)$ into a row vector $w_k^\top$. Every hinge in $M$ already has the orientation $i<j$, so its coefficient is added directly to the corresponding feature.

To specify the remaining row, collect the auxiliary coefficients of $R$ as
\begin{align}\label{eq:auxiliary-remainder}
R(x)=\sum_i p_i\relu(x_i)+\sum_i n_i\relu(-x_i)+\sum_{i<j}s_{ij}\relu(x_i-x_j).
\end{align}
Each retained unit is translation invariant by~\eqref{eq:translation-invariance}, so
\begin{align}\label{eq:remainder-translation}
R(x+t\one)=R(x)+t,\qquad \sum_i p_i=1.
\end{align}
The coefficient sum follows by setting $x=0$ and $t=1$. Now choose $t$ large enough that every $x_i+t$ is nonnegative. Substituting the auxiliary expansion into~\eqref{eq:remainder-translation} cancels $t$ and gives
\begin{align}\label{eq:compact-remainder}
R(x)=\sum_i p_ix_i+\sum_{i<j}s_{ij}\relu(x_i-x_j).
\end{align}
Substitute~\eqref{eq:recover-coordinates} into this expression and collect the coefficients of $\Phi(x)$ into $w_R$. Thus both $w_k$ and $w_R$ are specified by finite sums of the previously defined coefficients.

For any fixed ordering of $K$, the Caterpillar Net has second-layer rows and output weights
\begin{align}\label{eq:network-assembly}
W_2=\begin{pmatrix}(w_k^\top)_{k\in K}\\w_R^\top\\-w_R^\top\end{pmatrix},\qquad
w_3=\begin{pmatrix}(d_k)_{k\in K}\\1\\-1\end{pmatrix},\qquad b_1=b_2=b_3=0.
\end{align}
The first $|K|$ neurons output $U_k(x)$, and the last two output $\relu(R(x))$ and $\relu(-R(x))$. Thus
\begin{align}\label{eq:signed-remainder}
w_3^\top\relu(W_2\Phi(x))=\sum_{k\in K}d_kU_k(x)+\relu(R(x))-\relu(-R(x))=\maxn_N(x).
\end{align}
By~\eqref{eq:retained-count}, the second-layer width is at most
\begin{align}\label{eq:compilation-widths}
|K|+2\leq3^{N-2}C_N-N(N-1)+2=h_2.
\end{align}
Zero padding realizes the stated widths exactly.

For the growth rates, count the ordered choices of $m$ spine labels:
\begin{align}\label{eq:head-count-bound}
\binom Nm\frac{m!}{2}=\frac{N!}{2(N-m)!}\leq N^m\qquad(2\leq m\leq N-1).
\end{align}
Using $N-m\leq N$ and $m^{N-m-1}\leq N^{N-m-1}$ in each summand of~\eqref{eq:count} gives
\begin{align}\label{eq:cone-growth}
\begin{aligned}
C_N&\leq\sum_{m=2}^{N-1}N^m\,N\,N^{N-m-1}=(N-2)N^N\\
&\leq N^{N+1}\leq N^{4N}.
\end{aligned}
\end{align}
Thus no Stirling estimate is needed. Since $3^{N-2}\leq N^N$ for $N\geq3$, the widths satisfy
\begin{align}
h_2&\leq3^{N-2}(N-2)N^N+2\leq2N^{5N},\label{eq:second-growth}\\
h_1&\leq4N^2.\label{eq:first-growth}
\end{align}
Taking logarithms gives~\eqref{eq:asymptotics}.
\finishes{Theorem}{thm:main}
\end{Proof}

\paragraph[Formalization]{Formalization (\leandecl{MaxCaterpillar/Theorems.lean}{18}{maximum_width})}
Linked definitions, lemmas, and theorems are audited against exact types;
only \texttt{propext}, \texttt{Classical.choice}, and \texttt{Quot.sound}
are allowed. The repository contains the maximum construction and its proof.
The general CPWL results in the introduction are cited background results;
the equation links in the Wang--Sun theorem lead to its published source.
Claude and Codex assisted with formalization and manuscript preparation.

\section{Conclusion}
The first hidden-layer width is asymptotically sharp. Building on the extremal-graph approach of \citet{SafranDepth}, our separate Lean development extends the bound $h_1\geq\binom N2-\lfloor4N/3\rfloor$ of \citet[Theorem~2.13]{LuClosure}, for $N\geq4$, to networks with arbitrary affine skip connections. When every first-layer weight vector is proportional to some $e_i-e_j$, we sharpen it to $h_1\geq\binom N2-N$. We obtain no growing lower bound for $h_2$; finding smaller exact maximum representations and establishing such lower bounds remain open.

\end{document}